\documentclass[lettersize,journal]{IEEEtran}
\usepackage{amsmath,amsfonts}
\usepackage{algorithmic}
\usepackage{algorithm}
\usepackage{array}
\usepackage[caption=false,font=normalsize,labelfont=sf,textfont=sf]{subfig}
\usepackage{textcomp}
\usepackage{stfloats}
\usepackage{url}
\usepackage{verbatim}
\usepackage{graphicx}
\usepackage{cite}
\usepackage[caption=false,font=normalsize,labelfont=sf,textfont=sf]{subfig}
\usepackage{color}
\usepackage{amsthm, amssymb,bm}

\usepackage{mathtools}
\usepackage{array}       
\usepackage{booktabs}    

\begin{document}

\title{Cooperative Circumnavigation for Multi-Quadrotor Systems via Onboard Sensing}


\author{Xueming Liu$^1$, Lin Li$^1$, Xiang Zhou$^1$, Qingrui Zhang$^{1,*}$, and Tianjiang Hu$^{1,2,*}$
\thanks{This work is supported by the Key-Area Research and Development Program of Guangdong Province under Grant 2024B1111060004, and in part by the National Natural Science Foundation of China under Grant 62473390, the Basic and Applied Basic Research Foundation of Guangdong Province under Grant 2024A1515012408,  the Shenzhen Science and Technology Program JCYJ20220530145209021, and State Key Laboratory of Robotics and Systems (HIT) under Grant SKLRS-2024-KF-05.}
\thanks{$^1$School of Aeronautics and Astronautics, Sun Yat-sen University (Shenzhen Campus), Shenzhen, China.}
\thanks{$^1$School of Aeronautics and Astronautics, Sun Yat-sen University (Shenzhen Campus), Shenzhen, China.}
\thanks{$^2$School of Artificial Intelligence, Sun Yat‐senUniversity, Guangzhou, China}
\thanks{$^*$Corresponding authors (\{zhangqr9, hutj3\}@mail.sysu.edu.cn). }}



\maketitle

\begin{abstract}
A cooperative circumnavigation framework is proposed for multi-quadrotor systems to enclose and track a moving target without reliance on external localization systems. The distinct relationships between quadrotor-quadrotor and quadrotor-target interactions are evaluated using a heterogeneous perception strategy and corresponding state estimation algorithms. A modified Kalman filter is developed to fuse visual-inertial odometry with range measurements to enhance the accuracy of inter-quadrotor relative localization. An event-triggered distributed Kalman filter is designed to achieve robust target state estimation under visual occlusion by incorporating neighbor measurements and estimated inter-quadrotor relative positions. Using the estimation results, a cooperative circumnavigation controller is constructed, leveraging an oscillator-based autonomous formation flight strategy. We conduct extensive indoor and outdoor experiments to validate the efficiency of the proposed circumnavigation framework in occluded environments. Furthermore, a quadrotor failure experiment highlights the inherent fault tolerance property of the proposed framework, underscoring its potential for deployment in search-and-rescue operations.
\end{abstract}

\begin{IEEEkeywords}
Circumnavigation, multi-quadrotor system, target tracking, state estimation.
\end{IEEEkeywords}

\section{Introduction}



Circumnavigation is a target enclosing and tracking strategy where one or more collaborative agents encircle a target at a specified distance \cite{litimeinSurveyTechniquesCircular2021,deghatLocalizationCircumnavigationSlowly2014}. This approach has been widely studied in applications such as marine biology \cite{shinzakiMultiAUVSystemCooperative2013}, satellite formation flying \cite{liCooperativeCircumnavigationControl2020}, source seeking \cite{mooreSourceSeekingCollaborative2010}, and aerial motion capture \cite{lopez-nicolasAdaptiveMultirobotFormation2020}.

With advances in autonomous quadrotor technology, single-quadrotor circumnavigation of ground targets has been extensively explored \cite{suiAdaptiveBearingOnlyTarget2024,wangTargetTrackingCircumnavigation2024}. However, rapid progress in formation control has shown that multi-quadrotor systems offer greater flexibility and robustness for target tracking \cite{chungSurveyAerialSwarm2018}. These systems can address challenges like environmental occlusions \cite{changCrossDroneBinocularCoordination2020,zhouSwarmMicroFlying2022,ibenthalLocalizationPartiallyHidden2023,liuFormationControlEnclosing2025}, target localization accuracy enhancement \cite{hausmanCooperativeMultirobotControl2015,priceDeepNeuralNetworkBased2018,xuDistributedPseudolinearEstimation2017}, and many more \cite{zhangRobustNonlinearClose2021, GBZhu_T-Mech2024}. However, cooperative circumnavigation poses significant challenges when global positioning information is unavailable for either the quadrotors or the target, which commonly occurs in search-and-rescue operations. In such a case, existing approaches dependent on global positioning systems \cite{liFullyDistributedCooperative2021,zhangCompositeSystemTheorybased2021,zhangRobustGuidanceLaw2023} become infeasible under such conditions due to the absence of external localization infrastructure.

    \begin{figure}[htbp]
        \centering
        \includegraphics[width=\linewidth]{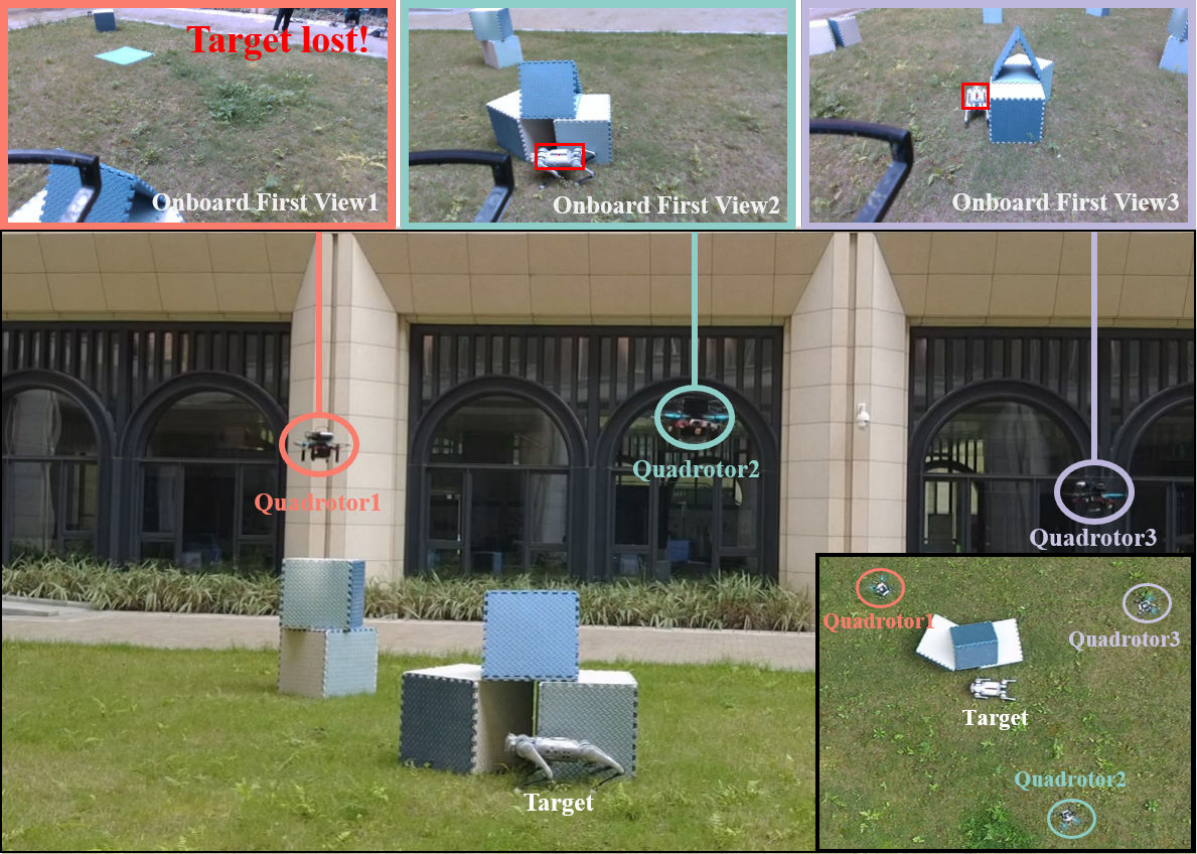}
        \caption{Three quadrotors performing cooperative circumnavigation around a ground target (a quadruped robot), with first-person camera perspectives from each agent. The left quadrotor loses target visibility due to occlusion, while the other two maintain observation. The bottom-right corner shows a top-down view.}
        \label{fig:show}
    \end{figure}

Under such conditions, quadrotors must simultaneously estimate their relative positions with respect to both a target and neighboring agents to enable successful cooperative circumnavigation. Though some studies assume that relative position information can be directly obtained \cite{chenFormationCircumnavigationUnmanned2019,liVGSwarmVisionBasedGene2023}, this assumption is restrictive and impractical for some multi-quadrotor applications. Acquiring the relative positions of neighboring agents requires complex sensor configurations and significant computational resources. For example, the omnidirectional vision method employing four cameras \cite{liVGSwarmVisionBasedGene2023} necessitates distinguishing all quadrotors and the target and may fail when a quadrotor is occluded by others.

To address these challenges, researchers have proposed simplified perception strategies based primarily on either bearing measurements \cite{zhengEnclosingTargetNonholonomic2015,liLocalizationCircumnavigationMultiple2018,zouCoordinateFreeDistributedLocalization2022} or range measurements \cite{hungCooperativeDistributedEstimation2022,liuMovingTargetCircumnavigationUsing2023,liuFormationControlMoving2023,zouLeaderFollowerCircumnavigation2024}. However, these approaches exhibit several practical limitations. First, they typically assume measurement homogeneity between quadrotor-quadrotor and quadrotor-target observations, despite the fundamental distinction between cooperative (quadrotor-quadrotor) and non-cooperative (quadrotor-target) sensing scenarios. This oversimplification often results in impractical sensor requirements. For instance, while visual bearing measurements to targets are generally obtainable, they become unreliable in multi-quadrotor systems subject to Non-Line-of-Sight (NLOS) conditions. Similarly, Ultra-Wide Band (UWB) technology can enable reliable inter-quadrotor ranging, but it remains a challenge to obtain accurate range measurements for non-cooperative targets. As a result, hybrid approaches that fuse both range and bearing measurements have been widely adopted for target position estimation, typically implemented using stereo vision systems \cite{nielsenRelativeMovingTarget2019} or monocular cameras combined with laser rangefinders \cite{ningRealtoSimtoRealApproachVisionBased2024}. Second, the existing studies primarily focus on unobstructed environments \cite{zhengEnclosingTargetNonholonomic2015,liLocalizationCircumnavigationMultiple2018,zouCoordinateFreeDistributedLocalization2022,hungCooperativeDistributedEstimation2022,liuMovingTargetCircumnavigationUsing2023,liuFormationControlMoving2023,zouLeaderFollowerCircumnavigation2024}, despite the prevalence of occlusions in real-world applications. Consequently, their applicability to cluttered or occluded scenarios remains limited. Furthermore, although the fault-tolerance advantages of multi-quadrotor systems have been widely recognized \cite{chungSurveyAerialSwarm2018,Zhang2015RAL}, few studies have developed practical frameworks capable of realizing this potential in dynamic or failure-prone environments.

\begin{figure}[tbp]
        \centering
        \includegraphics[width=\linewidth]{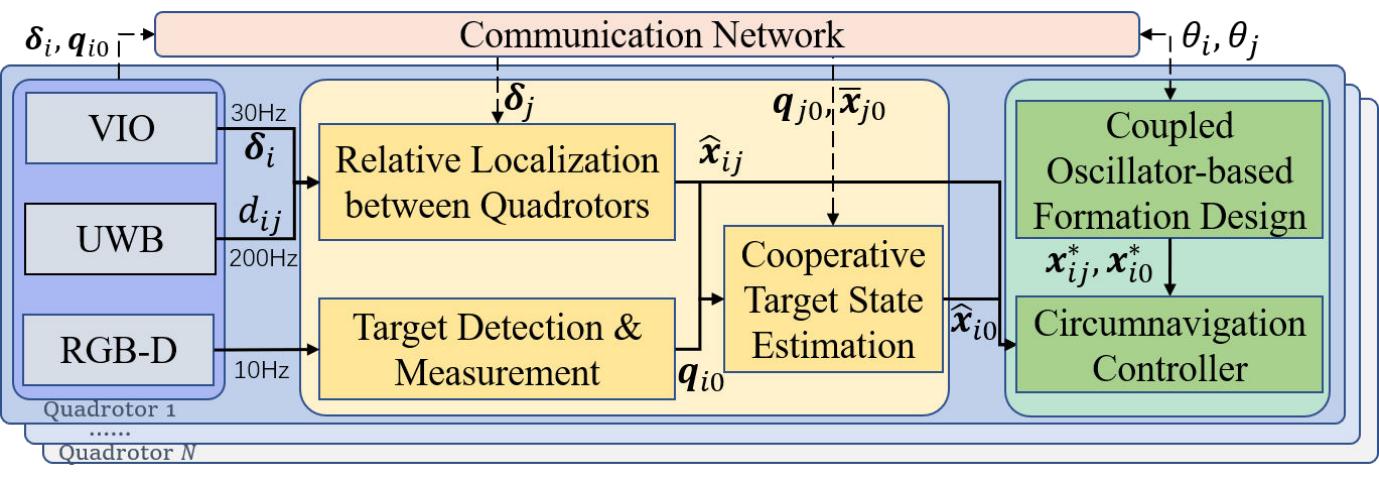}
        \caption{The proposed cooperative circumnavigation framework.}
        \label{fig:structure}
    \end{figure}

This paper addresses these challenges by proposing distinct perception and estimation strategies for quadrotor-quadrotor and quadrotor-target interactions, as illustrated in Fig.~\ref{fig:structure}. The proposed framework incorporates three key components: (1) a modified Kalman Filter (KF) that fuses UWB-based ranging and Visual-Inertial Odometry (VIO) data for inter-quadrotor localization; (2) a stereo vision-based target state estimation module with an event-triggered Distributed Kalman Filter (DKF) for handling occlusion conditions; and (3) a coupled oscillator-based formation control scheme \cite{liuFormationControlMoving2023} capable of autonomous adaptation to varying swarm sizes.


\section{RELATED WORKS AND CONTRIBUTIONS} \label{relative_work}
\subsection{Relative Localization in Multi-Agent Systems}
In recent years, UWB has been widely used in multi-quadrotor systems for relative localization estimation \cite{nguyenDistanceBasedCooperativeRelative2019,nguyenPersistentlyExcitedAdaptive2020,guoUltraWidebandOdometryBasedCooperative2020,cossetteRelativePositionEstimation2021,zhangAgileFormationControl2022,zhengUWBVIOFusionAccurate2022,fishbergMURPMultiAgentUltraWideband2024}, often in combination with odometry measurements from VIO or Simultaneous Localization and Mapping (SLAM). This widespread adoption is due to UWB’s dual capabilities in ranging and communication, as well as its effectiveness under NLOS conditions. Two main approaches are typically available in the literature to estimate high-dimensional relative positions from one-dimensional range measurements. The first one involves installing multiple UWB sensors on the quadrotor \cite{cossetteRelativePositionEstimation2021,zhengUWBVIOFusionAccurate2022,fishbergMURPMultiAgentUltraWideband2024}. However, due to the limited size of the quadrotor, this approach is impractical for long-range scenarios or small quadrotors. The second approach uses a single UWB sensor, requiring sufficient relative motion changes between quadrotors to ensure the convergence of the localization estimator \cite{nguyenDistanceBasedCooperativeRelative2019,nguyenPersistentlyExcitedAdaptive2020,guoUltraWidebandOdometryBasedCooperative2020}. Clearly, the second approach offers advantages in terms of sensor configuration and limited onboard resources. However, the studies in \cite{nguyenDistanceBasedCooperativeRelative2019,nguyenPersistentlyExcitedAdaptive2020} completely neglect measurement noise, while \cite{guoUltraWidebandOdometryBasedCooperative2020} accounts only for errors arising from relative motion. In practice, both UWB and relative motion measurements are affected by noise, which introduces bias in the estimation results. To mitigate the estimation bias, a modified KF algorithm is proposed to improve inter-quadrotor relative localization accuracy in this paper.

\subsection{Cooperative Target State Estimation}
Conventional methods for cooperative target state estimation are mainly based on the DKF framework with consensus strategies. The core idea of DKF is that each agent independently processes its direct measurements of the target while fusing indirect measurements from neighboring agents to reach a consensus estimate, thus preventing task failure caused by the loss of a single agent's direct measurements \cite{wangCooperativeTargetTracking2012}. However, a common assumption in these studies is that the positions of all agents \cite{olfati-saberCollaborativeTargetTracking2011,  hungCooperativeDistributedEstimation2022, xuDistributedPseudolinearEstimation2017}, or at least a subset \cite{wangCooperativeTargetTracking2012, liDistributedKalmanFilter2020, doostmohammadianDistributedEstimationApproach2022}, are known in a global reference frame, which is impractical in GPS-denied environments. To address this, this paper proposes a solution by integrating relative state estimation between quadrotors into the DKF framework. However, the quality of a quadrotor's direct measurements of the target is considered superior to indirect measurements, which combine neighbor measurements and relative state estimates between quadrotors. To account for this, an event-triggered strategy \cite{dingOverviewRecentAdvances2018, prielEventtriggeredConsensusKalman2023} is applied to cooperative target state estimation. In this strategy, indirect measurements from neighbors are used only when a quadrotor loses its direct measurements due to environmental occlusion; otherwise, direct measurements are used to update the estimate.

\subsection{Contributions}
In summary, the contributions of this work are threefold:
\begin{enumerate}
    \item A cooperative circumnavigation strategy is proposed for multi-quadrotor systems, relying exclusively on onboard sensing. Distinct perception strategies and corresponding state estimation methods are developed to address both inter-quadrotor and quadrotor-target interactions. Compared with conventional homogeneous perception approaches \cite{zhengEnclosingTargetNonholonomic2015,liLocalizationCircumnavigationMultiple2018,zouCoordinateFreeDistributedLocalization2022,hungCooperativeDistributedEstimation2022,liuMovingTargetCircumnavigationUsing2023,liuFormationControlMoving2023,zouLeaderFollowerCircumnavigation2024}, the proposed heterogeneous framework demonstrates practical performance.
    
    \item A modified KF algorithm is introduced, effectively reducing estimation bias compared to the classical KF and recursive least square algorithm \cite{nguyenPersistentlyExcitedAdaptive2020}. Additionally, an event-triggered DKF algorithm is presented, enabling robust target state estimation in visually obstructed environments through integration of relative state estimates among quadrotors.
    
    \item Comprehensive real-world experiments are conducted in both indoor and outdoor environments to validate the algorithm's robustness under occlusion conditions. A quadrotor failure scenario is designed and implemented, successfully verifying the system's inherent fault-tolerant capabilities.
\end{enumerate}

\section{PROBLEM FORMULATION} \label{problem_formulation}
\begin{figure}[tbp]
        \centering
        \includegraphics[width=0.9\linewidth]{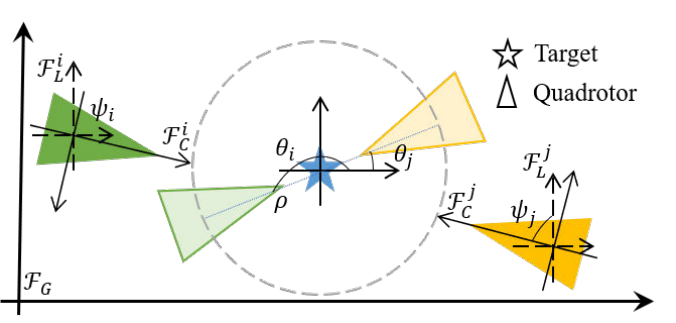}
        \caption{Illustration of the coordinate systems. Each quadrotor's local frame $\mathcal{F}_L^i$ maintains globally aligned orientation. The camera frame $\mathcal{F}_C^i$ is rigidly coupled to the quadrotor's yaw degree of freedom. The current quadrotor positions are denoted by dark triangles, while the desired positions (determined by the formation  parameter $\theta$ and $\rho$) are represented by light triangles.}
        \label{fig:frames}
    \end{figure}

Consider a multi-quadrotor system consisting of $N$ quadrotors collaborating to circumnavigate a ground target using only on-board sensors. In this work, each quadrotor can communicate with their neighbors to share information. Each quadrotor determines its ego-motion state within its local frame $\mathcal{F}_L^i$, using VIO. The local frames of all quadrotors are aligned to a common orientation relative to a global frame, $\mathcal{F}_G$, either by initializing their VIO systems in the same direction or by synchronizing all frames with the Earth's magnetic field via a compass. Furthermore, the quadrotors utilize UWB sensors to measure relative ranges between each other. The quadrotor is also equipped with a stereo camera for measuring the target's relative position. The stereo camera is rigidly mounted on the quadrotor with its own camera frame, $\mathcal{F}_C^i$. Consequently, the camera orientation is governed by the quadrotor's yaw control. In this work, it is assumed that all quadrotors fly at a constant altitude. Therefore, to simplify the problem, only the states of the target and the quadrotors in the $xy$-plane are considered.  The relationship between the coordinate systems is provided in Fig.~\ref{fig:frames}.

\subsection{Dynamic Model}
In the global frame $\mathcal{F}_G$, let $\boldsymbol{p}_{0,k}$ and $\boldsymbol{v}_{0,k} \in \mathbb{R}^2$ represent the position and velocity of the target, respectively, while $\boldsymbol{p}_{i,k}$ and $\boldsymbol{v}_{i,k} \in \mathbb{R}^2$ denote the position and velocity of quadrotor $i \in \{1, 2, \dots, N\}$, respectively. All states are measured in discrete-time manner with $t_k=k \Delta t$,  where $k\in \mathbb{N}$ denotes the time step and $\Delta t$ is the sampling period. The discrete double-integrator models for both the target and quadrotors are given by
    \begin{equation}\label{eq:dynamic_target}
        \boldsymbol{x}_{0,k+1} = \bm{\mathit{A}} \boldsymbol{x}_{0,k} + \boldsymbol{\omega}_{0,k}
    \end{equation}
    \begin{equation}\label{eq:dynamic_uav}
        \boldsymbol{x}_{i,k+1} = \bm{\mathit{A}} \boldsymbol{x}_{i,k} + \bm{\mathit{B}} \boldsymbol{u}_{i,k} + \boldsymbol{\omega}_{i,k}
    \end{equation}
where
      $$ \boldsymbol{x}_{0,k} = \begin{bmatrix} \boldsymbol{p}_{0,k} \\ \boldsymbol{v}_{0,k} \end{bmatrix}, \boldsymbol{x}_{i,k} = \begin{bmatrix} \boldsymbol{p}_{i,k} \\ \boldsymbol{v}_{i,k} \end{bmatrix}, \bm{\mathit{A}} = \begin{bmatrix} \bm{\mathit{I}} & \Delta t \bm{\mathit{I}} \\ \boldsymbol{0} & \bm{\mathit{I}} \end{bmatrix}, \bm{\mathit{B}}  = \begin{bmatrix} \boldsymbol{0} \\  \Delta t \bm{\mathit{I}} \end{bmatrix} $$
and, $ \boldsymbol{\omega}_{0,k}$ and $\boldsymbol{\omega}_{i,k}$ are the process noise, $\boldsymbol{u}_{i,k} \in \mathbb{R}^2$ is the control input of the quadrotors.

\subsection{Onboard Measurements}
1) \emph{Self-displacement Measurement via VIO:} VIO is a widely used onboard localization solution that offers good local accuracy but suffers from long-term drift \cite{xuOmniSwarmDecentralizedOmnidirectional2022}. At each time step $k$, the self-displacement $\boldsymbol{\delta}_{i,k}$ of the quadrotor $i$ is measured, given by
    \begin{equation}\label{eq:vio}
        \boldsymbol{\delta}_{i,k} = \boldsymbol{p}_{i,k} - \boldsymbol{p}_{i,k-1} + \boldsymbol{\mu}_{i,\boldsymbol{\delta},k}
    \end{equation}
where $\boldsymbol{\mu}_{i,\boldsymbol{\delta},k}$ is the measurement noise. It is important to emphasize that the position $\boldsymbol{p}_{i,k}$ of the quadrotor in the global coordinate frame is unknown and $\boldsymbol{\delta}_{i,k}$ is obtained directly through VIO. Quadrotor $i$ sends $\boldsymbol{\delta}_{i,k}$ to neighbouring quadrotors and receives $\boldsymbol{\delta}_{j,k}$ from neighbor $j$ via a network. Thus, the relative displacement between the quadrotor $i$ and $j$ can be calculated by
\begin{equation}\label{eq:relative_displacement}
        \begin{aligned}
            \boldsymbol{\delta}_{ij,k} &= \boldsymbol{\delta}_{i,k} - \boldsymbol{\delta}_{j,k} \\
                                       &= \boldsymbol{p}_{ij,k} - \boldsymbol{p}_{ij,k-1} + \boldsymbol{\mu}_{ij,\boldsymbol{\delta},k}
          \end{aligned} 
    \end{equation}
where $\boldsymbol{p}_{ij,k} = \boldsymbol{p}_{i,k} - \boldsymbol{p}_{j,k} $ is the relative position between quadrotor $i$ and $j$, and $\boldsymbol{\mu}_{ij,\boldsymbol{\delta},k} = \boldsymbol{\mu}_{i,\boldsymbol{\delta},k} - \boldsymbol{\mu}_{j,\boldsymbol{\delta},k}$ is the measurement noise. It should be noted that only relative measurements within the sampling time period $\Delta t$ are used, reducing the drift impact caused by long-term error accumulation in VIO.

2) \emph{Range Measurement via UWB:} A notable characteristic of UWB is its susceptibility to external interference, which can lead to significant outliers \cite{zhangAgileFormationControl2022,fishbergMURPMultiAgentUltraWideband2024,zhengUWBVIOFusionAccurate2022}. In light of this, this paper first preprocesses the UWB data to remove outliers and smooth it, as outlined in Algorithm \ref{alg:preprocessing_uwb}. Suppose that UWB measurements are taken at a time interval of $\Delta t_{uwb}$, and at time step $n\in \mathbb{N}$, the relative range to the neighbor, $d_{ij,n}^{uwb}$, is measured. The UWB sampling frequency is typically very high, reaching up to 200 Hz. Consequently, it is assumed that $\Delta t_{uwb} \ll \Delta t$. Let the standard deviation of the UWB range error be $\sigma_{uwb}^{*}$. When new data arrives, outliers are initially removed using the three-sigma rule (Line 3 in Algorithm \ref{alg:preprocessing_uwb}). Next, an Exponentially Weighted Mean (EWM) method is applied to smooth the data (Line 4 in Algorithm \ref{alg:preprocessing_uwb}). Finally, downsampling is performed, yielding the measurement $d_{ij,k}$, formally expressed as
    \begin{equation}\label{eq:uwb_distance}
            d_{ij,k} = \left \| \boldsymbol{p}_{i,k} - \boldsymbol{p}_{j,k} \right \| + \boldsymbol{\mu}_{ij,d,k}
    \end{equation}
where $\boldsymbol{\mu}_{ij,d,k}$ is the measurement noise.

\begin{algorithm}[H]
\caption{Preprocessing of UWB measurements} \label{alg:preprocessing_uwb}
\begin{algorithmic} [1]
\STATE \textbf{Init Parameters:} $\Delta t$, $\Delta t_{uwb}$, $\sigma_{uwb}^{*}$, $\beta$
\vspace{0.1cm}
\STATE \textbf{while} new measurement $d_{ij,n}^{uwb}$
\vspace{0.1cm}
\STATE \hspace{0.5cm}\textbf{if} $\left|d_{ij,n}^{uwb} - d_{ij,n-1}^{uwb}\right| \leq 3 \sigma_{uwb}$ \textbf{then}
\vspace{0.1cm}
\STATE \hspace{0.5cm}\hspace{0.5cm}$d_{ij,n}^{uwb} \gets \beta d_{ij,n-1}^{uwb} + (1-\beta)d_{ij,n}^{uwb}$
\vspace{0.1cm}
\STATE \hspace{0.5cm}\textbf{else} $d_{ij,n}^{uwb} \gets d_{ij,n-1}^{uwb}$
\vspace{0.1cm}
\STATE \hspace{0.5cm}\textbf{if} $n \Delta t_{uwb} = k \Delta t $ \textbf{then}
\vspace{0.1cm}
\STATE \hspace{0.5cm}\hspace{0.5cm}$d_{ij,k} \gets d_{ij,n}^{uwb}$

\end{algorithmic}
\end{algorithm}


3) \emph{Relative Position Measurement via Stereo Camera:} The stereo camera is capable of obtaining depth information, which enables the measurement of the relative position of the target. This paper employs YOLOv5s \cite{Jocher_YOLOv5_by_Ultralytics_2020} for target detection, achieving a detection frequency of 10 Hz. The algorithm outputs the target's pixel coordinates $(U,V)$ in the image. The depth of the target, $D$, can then be obtained from the depth image. The camera intrinsic matrix $\bm{\mathit{K}} \in \mathbb{R}^{3 \times 3}$ is assumed to be known. Hence, the relative position of the target in the camera frame $\mathcal{F}_C^i$, denoted as $\begin{bmatrix} X_C & Y_C & Z_C \end{bmatrix}^T$, is given by
    \begin{equation*} 
            \begin{bmatrix} X_C \\ Y_C \\ Z_C \end{bmatrix} = \bm{\mathit{R}}_C \cdot \bm{\mathit{K}}^{-1} \cdot \begin{bmatrix} U \\ V \\ 1 \end{bmatrix} \cdot D + \bm{\mathit{T}}_C
    \end{equation*}
where $\bm{\mathit{R}}_C \in \mathbb{R}^{3 \times 3}$ is the rotation matrix and $\bm{\mathit{T}}_C \in \mathbb{R}^{3 \times 1}$ is the translation vector, both of which depend on the camera's mounting pose. Meanwhile, the attitude of the quadrotor can be obtained through VIO. For simplicity, this paper considers only the yaw angle $\psi$. The relative position of the target in the local frame $\mathcal{F}_L^i$, denoted as $\begin{bmatrix} X_L & Y_L & Z_L \end{bmatrix}^T$, can be calculated by
    \begin{equation*} 
            \begin{bmatrix} X_L \\ Y_L \\ Z_L \end{bmatrix} = \begin{bmatrix} \cos{\psi} & -\sin{\psi} & 0 \\ \sin{\psi} & \cos{\psi} & 0 \\ 0 & 0 & 1  \end{bmatrix}^{-1} \begin{bmatrix} X_C \\ Y_C \\ Z_C \end{bmatrix}
    \end{equation*}
Consequently, the relative position of the target in $xy$-plane is 
    \begin{equation} \label{target_relative_pos}
            \boldsymbol{q}_{i0,k} \triangleq \begin{bmatrix} X_L \\ Y_L \end{bmatrix} = \boldsymbol{p}_{i,k} - \boldsymbol{p}_{0,k} + \boldsymbol{\mu}_{i0,\boldsymbol{q},k}
    \end{equation}
where $\boldsymbol{\mu}_{i0,\boldsymbol{q},k}$ is the measurement noise.

\section{Relative State Estimation} \label{estimation}

\subsection{Relative State Estimation between Quadrotors} \label{interuav}

According to \eqref{eq:dynamic_uav}, the relative dynamic between quadrotors is given as
    \begin{equation}\label{eq:relative_dynamic_uav}
        \boldsymbol{x}_{ij,k+1} = \bm{\mathit{A}} \boldsymbol{x}_{ij,k} + \bm{\mathit{B}} \boldsymbol{u}_{ij,k} + \boldsymbol{\omega}_{ij,k}
    \end{equation}
where $$\boldsymbol{x}_{ij,k} = \begin{bmatrix} \boldsymbol{p}_{ij,k} \\ \boldsymbol{v}_{ij,k} \end{bmatrix} = \begin{bmatrix} \boldsymbol{p}_{i,k} \\ \boldsymbol{v}_{i,k} \end{bmatrix} - \begin{bmatrix} \boldsymbol{p}_{j,k} \\ \boldsymbol{v}_{j,k} \end{bmatrix}, \boldsymbol{u}_{ij,k} = \boldsymbol{u}_{i,k} - \boldsymbol{u}_{j,k}$$ and $\boldsymbol{\omega}_{ij,k} \backsim \mathcal{N}(0,\bm{\mathit{Q}}_{ij})$ is a zero-mean Gaussian distribution with a covariance matrix $\bm{\mathit{Q}}_{ij}$, representing the process noise.


According to \cite{liuFormationControlMoving2023,liuFormationControlEnclosing2025}, the following quantities are defined $$y_{ij,k} \triangleq \frac{1}{2} \left[ d_{ij,k}^{2}- d_{ij,k-1}^{2}-\left\lVert \boldsymbol{\delta}_{ij,k} \right\rVert ^{2}  \right]$$  $$ \boldsymbol{\xi}_{ij,k} \triangleq \boldsymbol{\delta}_{ij,k} + \Delta t^2 \boldsymbol{u}_{ij,k-1} $$
Thus, let $\boldsymbol{z}_{ij,k} \triangleq  \begin{bmatrix} y_{ij,k} & \boldsymbol{\xi}_{ij,k}^T \end{bmatrix} ^T $.  The measurement model for the relative state estimator between quadrotors is given by
    \begin{equation}\label{eq:relative_measurement_uav}
        \boldsymbol{z}_{ij,k} = \bm{\mathit{H}}_{ij,k}^{*} \boldsymbol{x}_{ij,k} + \boldsymbol{\mu}_{ij,k}^{*}, \quad \bm{\mathit{H}}_{ij,k}^{*} = \begin{bmatrix} \boldsymbol{\delta}_{ij,k}^{*T} & \boldsymbol{0}_{1 \times 2} \\ \boldsymbol{0}_{2 \times 2} &  \Delta t \bm{\mathit{I}} \end{bmatrix}
    \end{equation}
For the measurement model \eqref{eq:relative_measurement_uav}, the following points require clarification. First,  $\boldsymbol{\mu}_{ij,k}^{*}$ denotes the measurement noise, which is determined by the noise in both $y_{ij,k}$ and $\boldsymbol{\xi}_{ij,k}$. It can be observed that $y_{ij,k}$ is a nonlinear combination of the range measurement $d_{ij,k}$ and the relative displacement measurement $\boldsymbol{\delta}_{ij,k}$. Recent studies \cite{fishbergMURPMultiAgentUltraWideband2024} suggest that UWB range measurement errors follow a long tail distribution rather than a Gaussian distribution, which complicates the analysis of the measurement error of $y_{ij,k}$. However, \cite{fishbergMURPMultiAgentUltraWideband2024} also indicates that assuming a Gaussian distribution can significantly improve system performance in many related works. Therefore, this paper also adopts the assumption that $\boldsymbol{\mu}_{ij,k}^{*} \backsim \mathcal{N}(0,\bm{\mathit{\Sigma}}_{ij}^{*})$ follows a zero-mean Gaussian distribution, where $\bm{\mathit{\Sigma}}_{ij}^{*}$ is the covariance matrix. Second, $\boldsymbol{\delta}_{ij,k}^{*} = \boldsymbol{p}_{ij,k} - \boldsymbol{p}_{ij,k-1}$ represents the true value of the relative displacement between the quadrotors. However, noisy measurements of the relative displacement $\boldsymbol{\delta}_{ij,k}$ are available, which corrupt the measurement matrix and consequently induce irreducible estimation bias.

To mitigate the estimation bias, a modified KF is proposed. First, the measurement matrix is decomposed into its deterministic component and noise term. This noise term is then combined with $\boldsymbol{\mu}_{ij,k}^{*}$ to form an augmented noise vector $\boldsymbol{\mu}_{ij,k}$.
    \begin{equation}\label{eq:relative_measurement_cov}
        \boldsymbol{\mu}_{ij,k} = - \text{tanh}(k) \cdot \begin{bmatrix} \boldsymbol{\mu}_{ij,\boldsymbol{\delta},k}^T\boldsymbol{\bar{p}}_{ij,k} \\ \boldsymbol{0}_{2 \times 1} \end{bmatrix} + \boldsymbol{\mu}_{ij,k}^{*}
    \end{equation}
where $\boldsymbol{\mu}_{ij,\boldsymbol{\delta},k} = \boldsymbol{\delta}_{ij,k}-\boldsymbol{\delta}_{ij,k}^{*}$ represents the measurement noise of the relative displacement, following a zero-mean Gaussian distribution with covariance matrix $\bm{\mathit{\Sigma}}_{ij,\boldsymbol{\delta}}$. Here, $\boldsymbol{\bar{p}}_{ij,k}$ denotes the prior relative position estimate of $\boldsymbol{ {p}}_{ij,k}$. The weighting function $\text{tanh}(k) = ({e^{ak}-e^{-ak}})/({e^{ak}+e^{-ak}})$, with $ 0 < a < 1$, regulates the noise-term weighting in accordance with the convergence of $\boldsymbol{\bar{p}}_{ij,k}$. By approximating $\boldsymbol{\bar{p}}_{ij,k}$ as deterministic and leveraging Gaussian distribution properties, $\boldsymbol{\mu}_{ij,k}$ preserves zero-mean Gaussian characteristics but exhibits a time-varying covariance matrix.
    \begin{equation}\label{eq:relative_measurement_covariance}
        \bm{\mathit{\Sigma}}_{ij,k} = \text{tanh}(k) \cdot \text{diag}\left[ \boldsymbol{\bar{p}}_{ij,k}^T \bm{\mathit{\Sigma}}_{ij,\boldsymbol{\delta}} \boldsymbol{\bar{p}}_{ij,k},0,0 \right] + \bm{\mathit{\Sigma}}_{ij}^{*}
    \end{equation}
Thus, the measurement model becomes
    \begin{equation}\label{eq:relative_measurement_uav_kfm}
        \boldsymbol{z}_{ij,k} \approx \bm{\mathit{H}}_{ij,k} \boldsymbol{x}_{ij,k} + \boldsymbol{\mu}_{ij,k}, \quad \boldsymbol{\mu}_{ij,k} \backsim \mathcal{N}(0,\bm{\mathit{\Sigma}}_{ij,k})
    \end{equation}

Consequently, based on the process model \eqref{eq:relative_dynamic_uav} and the measurement model \eqref{eq:relative_measurement_uav_kfm}, the following modified KF algorithm is derived to estimate the relative state between two quadrotors. Specifically, the prediction step is given by
    \begin{equation}\label{eq:predictKFM}
          \begin{aligned}
              & \boldsymbol{\bar{x}}_{ij,k} = \bm{\mathit{A}} \boldsymbol{\hat{x}}_{ij,k-1} + \bm{\mathit{B}} \boldsymbol{u}_{ij,k-1} \\
              & \bm{\mathit{P}}_{ij,k}^{-} = \bm{\mathit{A}} \bm{\mathit{P}}_{ij,k-1}^{+} \bm{\mathit{A}}^T + \bm{\mathit{Q}}_{ij}
          \end{aligned}
  \end{equation}
where $ \boldsymbol{\bar{x}}_{ij,k} = \left[ \begin{matrix} \boldsymbol{\bar{p}}_{ij,k}^T & \boldsymbol{\bar{v}}_{ij,k}^T \end{matrix} \right]^T$ and $ \boldsymbol{\hat{x}}_{ij,k} = \left[ \begin{matrix} \boldsymbol{\hat{p}}_{ij,k}^T & \boldsymbol{\hat{v}}_{ij,k}^T \end{matrix} \right]^T$ are the prior and posterior state estimates, respectively. In addition, $ \bm{\mathit{P}}_{ij,k}^{-}$ and $\bm{\mathit{P}}_{ij,k}^{+}$ are the prior and posterior error covariances of the estimate, respectively. Thus, the correction step is 
    \begin{equation}\label{eq:correctKFM}
          \begin{aligned}
              \bm{\mathit{K}}_{ij,k} &= \bm{\mathit{P}}_{ij,k}^{-} \bm{\mathit{H}}_{ij,k}^{T} \left( \bm{\mathit{H}}_{ij,k} \bm{\mathit{P}}_{ij,k}^{-} \bm{\mathit{H}}_{ij,k}^{T} + \bm{\mathit{\Sigma}}_{ij,k} \right)^{-1}  \\
              \boldsymbol{\hat{x}}_{ij,k} &= \boldsymbol{\bar{x}}_{ij,k} + \bm{\mathit{K}}_{ij,k} \left( \boldsymbol{z}_{ij,k} - \bm{\mathit{H}}_{ij,k} \boldsymbol{\bar{x}}_{ij,k} \right) \\
              \bm{\mathit{P}}_{ij,k}^{+} &= \left( \bm{\mathit{I}} - \bm{\mathit{K}}_{ij,k} \bm{\mathit{H}}_{ij,k} \right) \bm{\mathit{P}}_{ij,k}^{-} \left( \bm{\mathit{I}} - \bm{\mathit{K}}_{ij,k} \bm{\mathit{H}}_{ij,k} \right)^T \\ &+ \bm{\mathit{K}}_{ij,k} \bm{\mathit{\Sigma}}_{ij,k} \bm{\mathit{K}}_{ij,k}^{T}
          \end{aligned}
  \end{equation}
In this paper, the noise covariance matrix $\bm{\mathit{\Sigma}}_{ij,k}$ is adjusted based on the estimation of $\boldsymbol{p}_{ij,k}$, thereby reducing the estimate bias compared to classical KF.

\subsection{Cooperative Target State Estimation}
According to \eqref{eq:dynamic_target} and \eqref{eq:dynamic_uav}, the relative dynamics between the quadrotor and the target are given by
    \begin{equation}\label{eq:relative_dynamic_uav_target}
        \boldsymbol{x}_{i0,k+1} = \bm{\mathit{A}} \boldsymbol{x}_{i0,k} + \bm{\mathit{B}} \boldsymbol{u}_{i0,k} + \boldsymbol{\omega}_{i0,k}
    \end{equation}
where $\boldsymbol{x}_{i0,k} = \boldsymbol{x}_{i,k} - \boldsymbol{x}_{0,k}$, $\boldsymbol{u}_{i0,k} = \boldsymbol{u}_{i,k}$, and $\boldsymbol{\omega}_{i0,k} \backsim \mathcal{N}(0,\bm{\mathit{Q}}_{i0})$ represent the process noise, which follows a zero-mean Gaussian distribution with covariance matrix $\bm{\mathit{Q}}_{i0}$.

Considering an environment with obstacles, we use $\mathcal{O}_{k}$ to denote the set of quadrotors that can directly observe the target through vision without obstruction. It is assumed that at least one quadrotor can observe the target, \emph{i.e.} $\mathcal{O}_{k} \neq \emptyset$. If $i\in \mathcal{O}_{k}$, the direct measurement of the target is used, which yields
    \begin{equation}\label{eq:direct_measure}
        \boldsymbol{z}_{i0,k}^{i} \triangleq \boldsymbol{q}_{i0,k} = \bm{\mathit{C}} \boldsymbol{x}_{i0,k} + \boldsymbol{\mu}_{i0,\boldsymbol{q},k}^{i}, \quad \bm{\mathit{C}} = \begin{bmatrix} \bm{\mathit{I}} & \boldsymbol{0} \end{bmatrix}
    \end{equation}
where $\boldsymbol{\mu}_{i0,\boldsymbol{q},k}^{i} \backsim \mathcal{N}(0,\bm{\mathit{\Sigma}}_{i0,\boldsymbol{q}}^{i})$ follows a zero-mean Gaussian distribution with covariance matrix $\bm{\mathit{\Sigma}}_{i0,\boldsymbol{q}}^{i}$. Otherwise, the indirect measurement of the target is used, given by
    \begin{equation}\label{eq:indirect_measure}
              \boldsymbol{z}_{i0,k}^{j} \triangleq \boldsymbol{q}_{j0,k} + \boldsymbol{\hat{p}}_{ij,k} = \bm{\mathit{C}} \boldsymbol{x}_{i0,k} + \boldsymbol{\mu}_{i0,\boldsymbol{q},k}^{j}
    \end{equation}
where $\boldsymbol{\mu}_{i0,\boldsymbol{q},k}^{j}$ represents the measurement noise, assumed to follow a zero-mean Gaussian distribution with covariance matrix $\bm{\mathit{\Sigma}}_{i0,\boldsymbol{q},k}^{j} = \bm{\mathit{\Sigma}}_{j0,\boldsymbol{q}}^{j} + \bm{\mathit{C}} \bm{\mathit{P}}_{ij,k}^{+} \bm{\mathit{C}}^T$. Since $\boldsymbol{\hat{p}}_{ij,k}$ contains errors, it is considered that direct measurements $\boldsymbol{z}_{i0,k}^{i}$ have higher quality compared to indirect measurement $\boldsymbol{z}_{i0,k}^{j}$. Therefore, an event-triggered measurement model is proposed as follows
    \begin{equation}\label{eq:measrement_model_target}
              \boldsymbol{z}_{i0,k} = \bm{\mathit{C}} \boldsymbol{x}_{i0,k} + \boldsymbol{\mu}_{i0,\boldsymbol{q},k}, \quad \boldsymbol{\mu}_{i0,\boldsymbol{q},k} \backsim \mathcal{N}(0,\bm{\mathit{\Sigma}}_{i0,\boldsymbol{q},k})
    \end{equation}
where $$\boldsymbol{z}_{i0,k} = \begin{cases} \boldsymbol{z}_{i0,k}^{i} & \text{if } i\in \mathcal{O}_{k} \\ \dfrac{1}{|\mathcal{O}_{k}|} \sum\limits_{j\in \mathcal{O}_{k}} \boldsymbol{z}_{i0,k}^{j} & \text{if } i\notin \mathcal{O}_{k} \end{cases}$$
Additionally, the corresponding covariance matrix is given by $$\bm{\mathit{\Sigma}}_{i0,\boldsymbol{q},k} = \begin{cases} \bm{\mathit{\Sigma}}_{i0,\boldsymbol{q}}^{i} & \text{if } i\in \mathcal{O}_{k} \\ \dfrac{1}{|\mathcal{O}_{k}|^2} \sum\limits_{j\in \mathcal{O}_{k}} \bm{\mathit{\Sigma}}_{i0,\boldsymbol{q},k}^{j} & \text{if } i\notin \mathcal{O}_{k} \end{cases}$$
In other words, the neighbor's measurement data is only used to assist estimation when the target is not visible due to obstruction.

Denote the prior and posterior estimates of $\boldsymbol{x}_{i0,k}$ for quadrotor $i$ as $\boldsymbol{\bar{x}}_{i0,k}^{i}$ and $\boldsymbol{\hat{x}}_{i0,k}^{i}$, respectively. Then, the prediction step of DKF is given as
    \begin{equation}\label{eq:predictDKF}
          \begin{aligned}
              \boldsymbol{\bar{x}}_{i0,k}^{i} &= \bm{\mathit{A}} \boldsymbol{\hat{x}}_{i0,k-1}^{i} + \bm{\mathit{B}} \boldsymbol{u}_{i0,k-1} \\
              \bm{\mathit{P}}_{i0,k}^{-} &= \bm{\mathit{A}} \bm{\mathit{P}}_{i0,k-1}^{+} \bm{\mathit{A}}^T + \bm{\mathit{Q}}_{i0}
          \end{aligned}
  \end{equation}
where $\bm{\mathit{P}}_{i0,k}^{-}$ and $\bm{\mathit{P}}_{i0,k}^{+}$ are the prior and posterior estimate error covariances, respectively. Meanwhile, let the indirect estimate of $\boldsymbol{x}_{i0,k}$ from quadrotor $j$ be denoted as $\boldsymbol{\bar{x}}_{i0,k}^{j}$ with
    \begin{equation}\label{eq:indirect_estimate}
              \boldsymbol{\bar{x}}_{i0,k}^{j} = \boldsymbol{\bar{x}}_{j0,k}^{j} + \boldsymbol{\hat{x}}_{ij,k}
  \end{equation}
with the associated estimate error covariance matrix given by $\bm{\mathit{P}}_{i0,k}^{j} = \bm{\mathit{P}}_{j0,k}^{-} + \bm{\mathit{P}}_{ij,k}^{+}$. Thus, following the classic DKF paradigm outlined in \cite{olfati-saberCollaborativeTargetTracking2011, liDistributedKalmanFilter2020, wangCooperativeTargetTracking2012}, which is expressed as \emph{``posterior = prior + innovation + consensus"}, the update step for the algorithm is given by
    \begin{equation}\label{eq:updateDKF}
        \begin{aligned}
          &\left(\bm{P}_{i0,k}^{+}\right)^{-1} \!\!\!\!=\! \left(\bm{P}_{i0,k}^{-}\right)^{-1} \!\!\!\!+\! \bm{C}^T\bm{\Sigma}_{i0,\boldsymbol{q},k}^{-1}\bm{C} \!+\! \!\sum_{\mathclap{j\in\mathbf{N}_i}}\! \left(\bm{P}_{i0,k}^{j}\right)^{-1} \!\!\! \\
          &\boldsymbol{\hat{x}}_{i0,k}^{i} \!=\! \boldsymbol{\bar{x}}_{i0,k}^{i} \!+\! \bm{\mathit{P}}_{i0,k}^{+} \bm{\mathit{C}}^T \bm{\mathit{\Sigma}}_{i0,\boldsymbol{q},k}^{-1} \left( \boldsymbol{z}_{i0,k} \!-\! \bm{\mathit{C}} \boldsymbol{\bar{x}}_{i0,k}^{i}   \right) \\
          & \qquad \qquad + \epsilon \bm{\mathit{P}}_{i0,k}^{+} \sum_{j\in \mathbf{N}_i} \left( \bm{\mathit{P}}_{i0,k}^{j} \right)^{-1} \!\! \left( \boldsymbol{\bar{x}}_{i0,k}^{j} \!\!-\! \boldsymbol{\bar{x}}_{i0,k}^{i} \right) \\
        \end{aligned}
  \end{equation}
where $ 0 < \epsilon < 1 $ is the consensus gain, and $\mathbf{N}_{i}$ denotes the set of neighbors of quadrotor $i$.
      
\section{Circumnavigation Control}
Leveraging the differential flatness property of quadrotors \cite{mellingerMinimumSnapTrajectory2011}, circumnavigation control is applied separately to the position and yaw angle. Specifically, formation control ensures the quadrotors are evenly distributed along a circle centered at the target, while yaw angle control orients the onboard camera toward the target.

\subsection{Formation Control}
In the cooperative circumnavigation task, the quadrotors are expected to orbit around the target on a circle with a radius of $\rho$. Therefore, the desired relative position $\boldsymbol{p}_{i0,k}^{*}$ and the relative velocity $\boldsymbol{v}_{i0,k}^{*}$ between quadrotor $i$ and the target are 
    \begin{equation}\label{eq:desired_pos_i0}
      \boldsymbol{p}_{i0,k}^{*} = \rho  \left[ \begin{array}{cc} \cos{\theta_{i,k}} \\  \sin{\theta_{i,k}} \\ \end{array} \right], \boldsymbol{v}_{i0,k}^{*} = \frac{1}{\Delta t} (\bm{\mathit{R}}_{\Delta \theta} - I ) \boldsymbol{p}_{i,k}^{*}
    \end{equation}
where $\theta_{i,k}$ denotes the angle of quadrotor $i$ relative to the target in the global frame $\mathcal{F}_G$ (see Fig.~\ref{fig:frames}), and $ \bm{\mathit{R}}_{\Delta \theta} \triangleq \begin{bmatrix} \cos \Delta \theta & - \sin \Delta \theta \\ \sin \Delta \theta & \cos \Delta \theta \end{bmatrix} $ is the rotation matrix. Here, $\Delta \theta$ represents the phase shift at each sampling interval, which determines the orbiting speed. Furthermore, the desired relative position $\boldsymbol{p}_{ij,k}^{*}$ and the relative velocity $\boldsymbol{v}_{ij,k}^{*}$ between quadrotor $i$ and $j$ are given as
    \begin{equation}\label{eq:desired_pos_ij}
      \boldsymbol{p}_{ij,k}^{*} = \boldsymbol{p}_{i0,k}^{*} - \boldsymbol{p}_{j0,k}^{*}, \quad \boldsymbol{v}_{ij,k}^{*} = \boldsymbol{v}_{i0,k}^{*} - \boldsymbol{v}_{j0,k}^{*}
    \end{equation}
To enable the quadrotors to autonomously negotiate and assign relative positions to the target, the method based on coupled oscillators from \cite{liuFormationControlMoving2023} is adopted by
    \begin{equation*}\label{eq:oscillator}
      \begin{split}
          \theta_{i,k+1} &= \theta_{i,k} + \Delta \theta + \Delta t\sum_{j = 1}^{N} \sum_{l = 1}^{N}  \frac{G_{l}}{lN}\sin{\left(l\left[\theta_{i,k}-\theta_{j,k}\right]\right)}
      \end{split}
    \end{equation*}
where $G_{l} \in \mathbb{R}$ (for $l=\{ 1,...,N \}$) denotes the coupling gains.

Based on the estimated and desired relative states, the cooperative circumnavigation controller is designed as
    \begin{equation}\label{eq:formationtrackingcontrol}
           \begin{aligned}
            &\boldsymbol{u}_{i,k} \!=\! \text{s}(U_1,\boldsymbol{u}_{i,k}^{1}) \!+\! \text{s}(U_2,\boldsymbol{u}_{i,k}^{2}) \!+\! \frac{1}{\Delta t} (\bm{\mathit{R}}_{\Delta \theta} \!-\! I ) \boldsymbol{v}_{i0,k}^{*}
            \end{aligned}
    \end{equation}
where the function $\text{s}(U,\boldsymbol{u}) = \left( \min \{U,\| \boldsymbol{u}\|\}/\|\boldsymbol{u}\| \right) \boldsymbol{u}$ limits the control inputs, with bounds $U_1,U_2>0$. The control terms ${u}_{i,k}^{1}$ and ${u}_{i,k}^{2}$ are given by
    \begin{equation*}
           \begin{aligned}
            &\boldsymbol{u}_{i,k}^{1} \! = \!- K_{p}\!\!\!\! \sum_{j\in\mathbf{N}_i\cup 0} \!\!\!\!\left(\boldsymbol{\hat{p}}_{ij,k}\!-\!\boldsymbol{p}_{ij,k}^*\right)\! - \!K_{v} \!\!\!\!\sum_{j\in\mathbf{N}_i\cup 0}\!\!\!\! \left(\boldsymbol{\hat{v}}_{ij,k}\!-\!\boldsymbol{v}_{ij,k}^*\right)\!\!\!\!\!  \\
            &\boldsymbol{u}_{i,k}^{2} \!= \!- K_{\rho} \left(\|\boldsymbol{\hat{p}}_{i0,k}\| - \rho \right) \boldsymbol{\hat{p}}_{i0,k} \\
            \end{aligned}
    \end{equation*}
with positive definite gains $K_{p},K_{v},K_{\rho}>0$.

\subsection{Yaw Control}
This subsection presents a yaw control law to ensure camera orientation toward the target. First, the dynamics of the yaw angle in the discrete model is given by $$\psi_{i,k+1} = \psi_{i,k} + u_{i,k}^{\psi}$$

Using the estimated target relative position $\boldsymbol{\hat{p}}_{i0,k} \triangleq \begin{bmatrix} \hat{p}_{i0,k}^{x} & \hat{p}_{i0,k}^{y} \end{bmatrix}^T $, the relative orientation is computed as $\hat{\psi}_{i,k} = \textit{atan2}(-\hat{p}_{i0,k}^{y}, -\hat{p}_{i0,k}^{x})$, where $\textit{atan2}(\cdot)$ ensures correct quadrant determination. Combining this with VIO-measured current yaw ${\psi}_{i,k}$, the control law is
    \begin{equation}\label{eq:yawcontrol}
           \begin{aligned}
            &u_{i,k}^{\psi} \!=\! - K_{\psi}({\psi}_{i,k} - \hat{\psi}_{i,k}), \quad K_{\psi} > 0
            \end{aligned}
    \end{equation}

\section{EXPERIMENTS} \label{experimental_results}
This section presents experimental validation of the proposed cooperative circumnavigation algorithm under challenging operational conditions, specifically environmental occlusions and simulated failure of a quadrotor during mission execution.

In the experiments, each quadrotor is equipped with 1) a PX4 Autopilot for low-level flight control; 2) an Intel NUC11TNH onboard computer for executing all algorithms through Robot Operating System (ROS); 3) an Intel RealSense T265 camera for VIO; 4) a NoopLoop UWB module for inter-agent ranging; and 5) an Intel RealSense D435 depth camera for target relative position measurement. The ground target is a Unitree Go1 quadruped robot, which is manually controlled via remote operation.

Experimental validation involved both indoor and outdoor testing. Indoor evaluations assessed the relative localization algorithm by comparing onboard estimates with high-precision ground truth measurements from an FZMotion\footnote{\url{https://www.lusterinc.com/FZMotion-Baidu/}} motion capture system. Subsequently, outdoor experiments evaluated the cooperative circumnavigation performance without ground-truth comparison, as external positioning systems were unavailable.

\subsection{Indoor - Relative Localization Between Quadrotors}
    \begin{figure}[tbp]
        \centering
        \includegraphics[width=\linewidth]{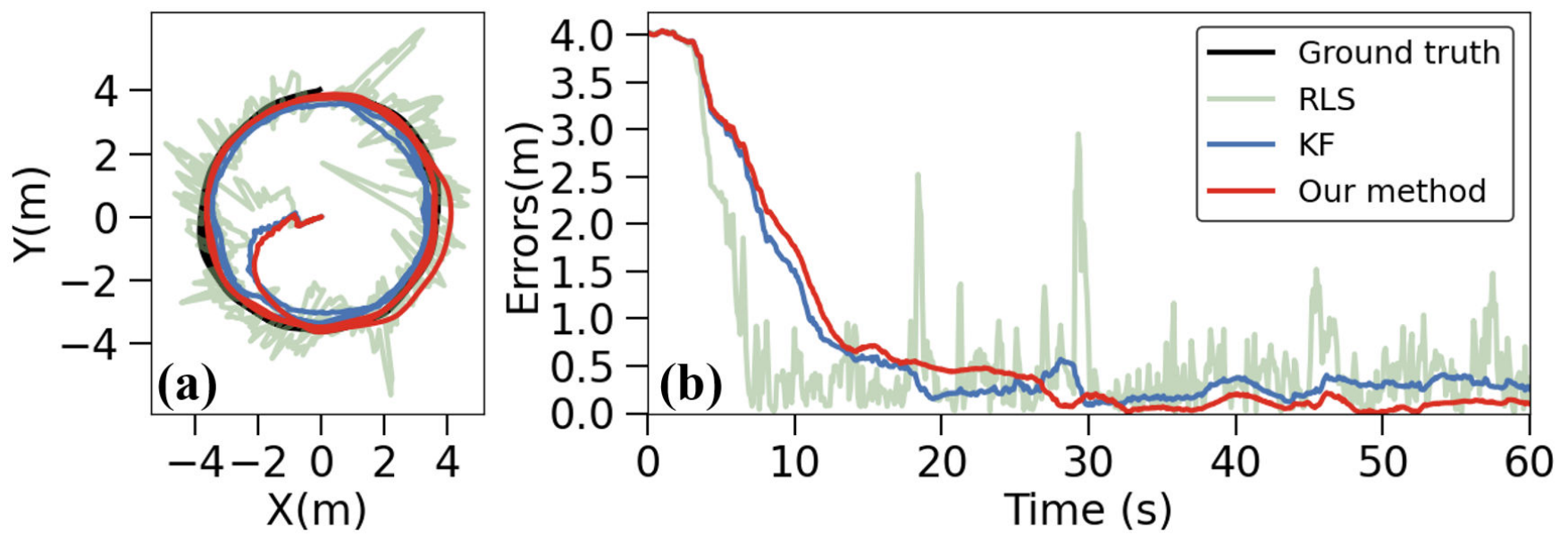}
        \caption{Relative localization results inter-quadrotors. (a) Estimated results $\boldsymbol{\hat{p}}_{ij,k}$ in the $xy$-plane. Black lines represents the ground truth results from a motion-capture system. (b) Inter-quadrotor relative position estimation error.}
        \label{fig:interuaverror}
    \end{figure}
Building upon the framework established in Section~\ref{interuav}, we first assess the inter-quadrotor relative localization performance. Our modified KF is benchmarked against both recursive least squares (RLS) estimator \cite{nguyenPersistentlyExcitedAdaptive2020} and classical KF implementations. The classical KF utilizes \eqref{eq:relative_measurement_uav} as its measurement equation, approximating true values $\boldsymbol{\delta}_{ij,k}^{*}$ with direct measurements $\boldsymbol{\delta}_{ij,k}$ while neglecting measurement noise.



    \begin{table}[htbp]
      \centering
      \caption{RMSE of inter-quadrotor relative localization}     
      \label{tab:interRMSE}
      \setlength{\tabcolsep}{12pt}
      \begin{tabular}{lccc}
        \hline 
        \textbf{Algorithm} & RLS & KF & Our method  \\
        \hline
        \textbf{RMSE}(m) & 0.63 & 0.43 & \textbf{0.22}  \\
        \hline
      \end{tabular}
    \end{table}

For performance comparison, two quadrotors fly along a $2m$ radius circular trajectory centered at the origin, completing each circle in 30 seconds. The quadrotors maintain symmetric positions about the center point. The noise covariance matrices are configured as $\bm{\mathit{Q}}_{ij}=0.0002\bm{\mathit{I}}$, $\bm{\mathit{\Sigma}}_{ij}^{*}=\textit{diag}(0.025,0.002,0.002)$, and $\bm{\mathit{\Sigma}}_{ij,\boldsymbol{\delta}}=\textit{diag}(0.002,0.002)$, with initial conditions $\boldsymbol{\hat{x}}_{ij,0}=\boldsymbol{0}$. Statistical results from 10 experimental trials show lower root mean square error (RMSE) values for our algorithm (Table~\ref{tab:interRMSE}), demonstrating effective bias reduction. Fig.~\ref{fig:interuaverror} (a) demonstrates the convergence of relative position estimates $\boldsymbol{\hat{p}}_{ij,k}$ in the $xy$-plane. Due to measurement matrix noise, the classical KF produces conservative estimates that consistently fall inside the ground truth trajectory obtained from the motion capture system. In contrast, our algorithm reduces estimation errors by separating the noise components in the measurement matrix and reconstructing the measurement noise covariance matrix. Fig.~\ref{fig:interuaverror} (b) illustrates the convergence characteristics of inter-quadrotor relative localization, where the estimation error is quantified by the Euclidean norm $e_{ij,k} = \|\boldsymbol{\hat{p}}_{ij,k} - \boldsymbol{{p}}_{ij,k} \|$. While the RLS method achieves faster initial convergence, it exhibits higher-magnitude and more volatile estimation errors compared to the KF-based approaches. In practical mission scenarios where quadrotor initial positions can be approximately estimated, proper initialization of the KF estimator will significantly improve its convergence rate.

    \begin{figure}[tbp]
        \centering
        \includegraphics[width=\linewidth]{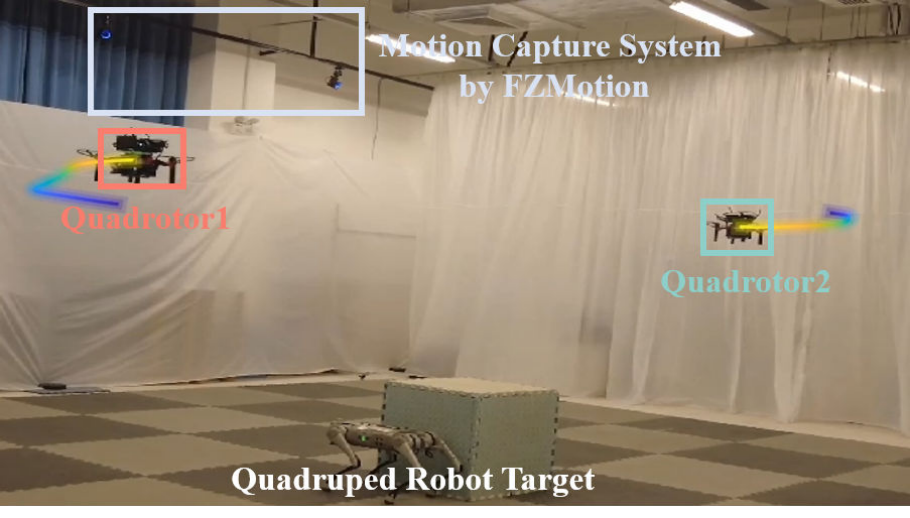}
        \caption{Two quadrotors performing cooperative circumnavigation around a stationary target in an indoor environment, with intermittent visual occlusion caused by a box during the mission.}
        \label{fig:indoorenv}
    \end{figure}

\subsection{Indoor - Cooperative Target Localization with Occlusions}
This subsection evaluates the event-triggered DKF algorithm's performance in occlusion scenarios. Due to limited indoor space, the experimental configuration utilizes two quadrotors performing circumnavigation around a stationary target, as shown in Fig.~\ref{fig:indoorenv}. 

\begin{figure}[tbp]
        \centering
        \includegraphics[width=\linewidth]{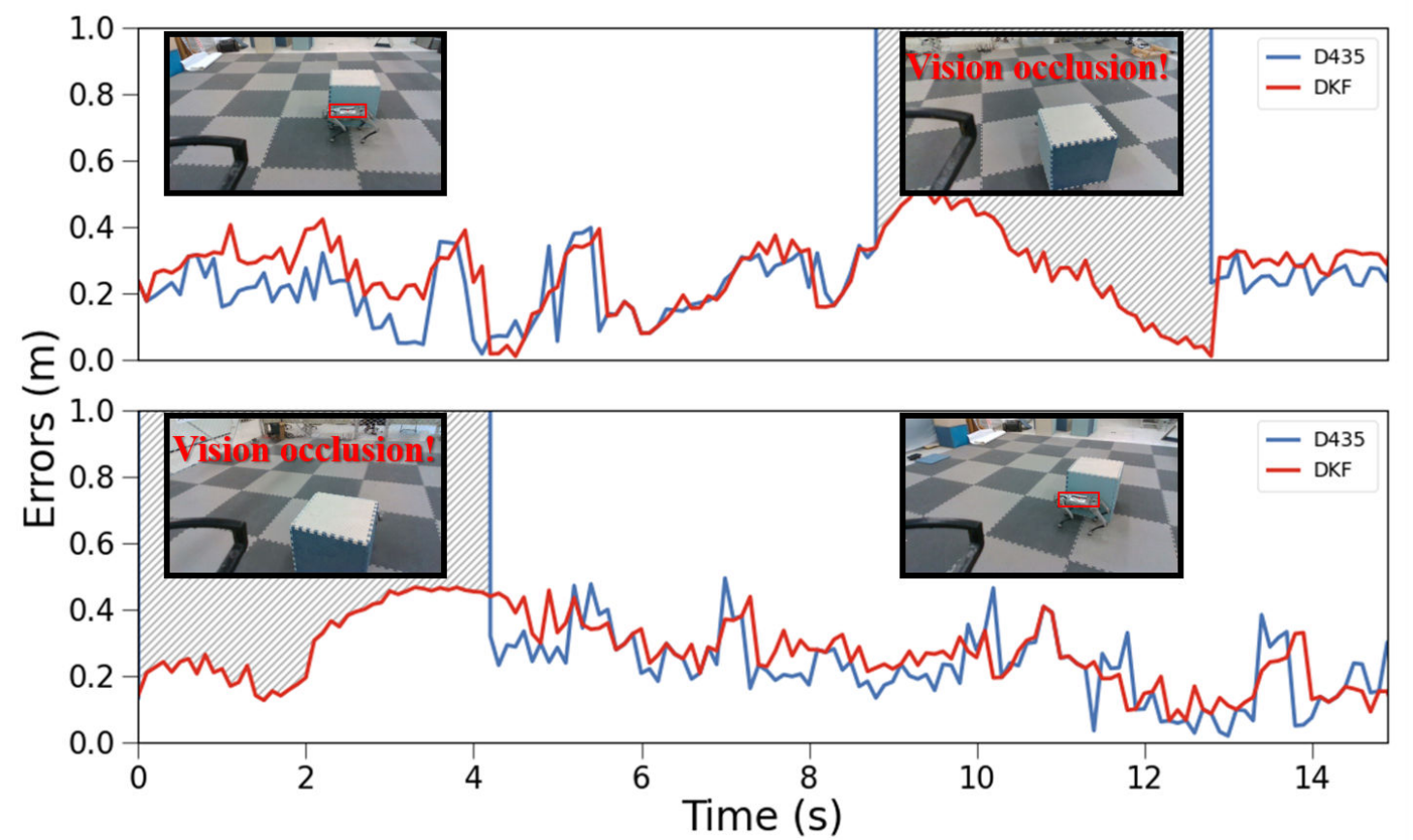}
        \caption{Measurement and estimation errors of the relative position with respect to the target for both quadrotors. The gray shaded area indicates the occluded region, with corresponding first-person views from the quadrotors provided. When one quadrotor loses target visibility, continuous estimation is maintained through cooperative information sharing between the two quadrotors.}
        \label{fig:indoordkf}
    \end{figure}

In Fig.~\ref{fig:indoordkf}, the relative position measurements obtained from the D435 camera and the estimation errors calculated by the DKF algorithm for each quadrotor with respect to the target are presented. Specifically, the measurement error is defined as $e_{i,k}^{m} = \|\boldsymbol{q}_{i0,k} - \boldsymbol{{p}}_{i0,k} \|$, while the estimation error is defined as $e_{i,k}^{e} = \|\boldsymbol{\hat{p}}_{i0,k} - \boldsymbol{{p}}_{i0,k} \|$. The cooperative system demonstrates robust performance during occlusion events: when one quadrotor loses target visibility due to occlusion, the other maintains observation. By fusing this maintained measurement with inter-quadrotor relative position estimates, continuous target tracking is achieved through indirect measurement. Our approach effectively compensates for individual sensor limitations while preserving estimation accuracy.

\subsection{Outdoor - Cooperative circumnavigation}
In the outdoor experiments, three quadrotors autonomously circumnavigate a moving Unitree Go1 quadruped robot, which is remotely controlled by an operator. Given the lack of precise outdoor positioning systems, these experiments primarily serve to validate the robustness of our algorithm in complex environments. 

\begin{figure}[tbp]
        \centering
        \includegraphics[width=\linewidth]{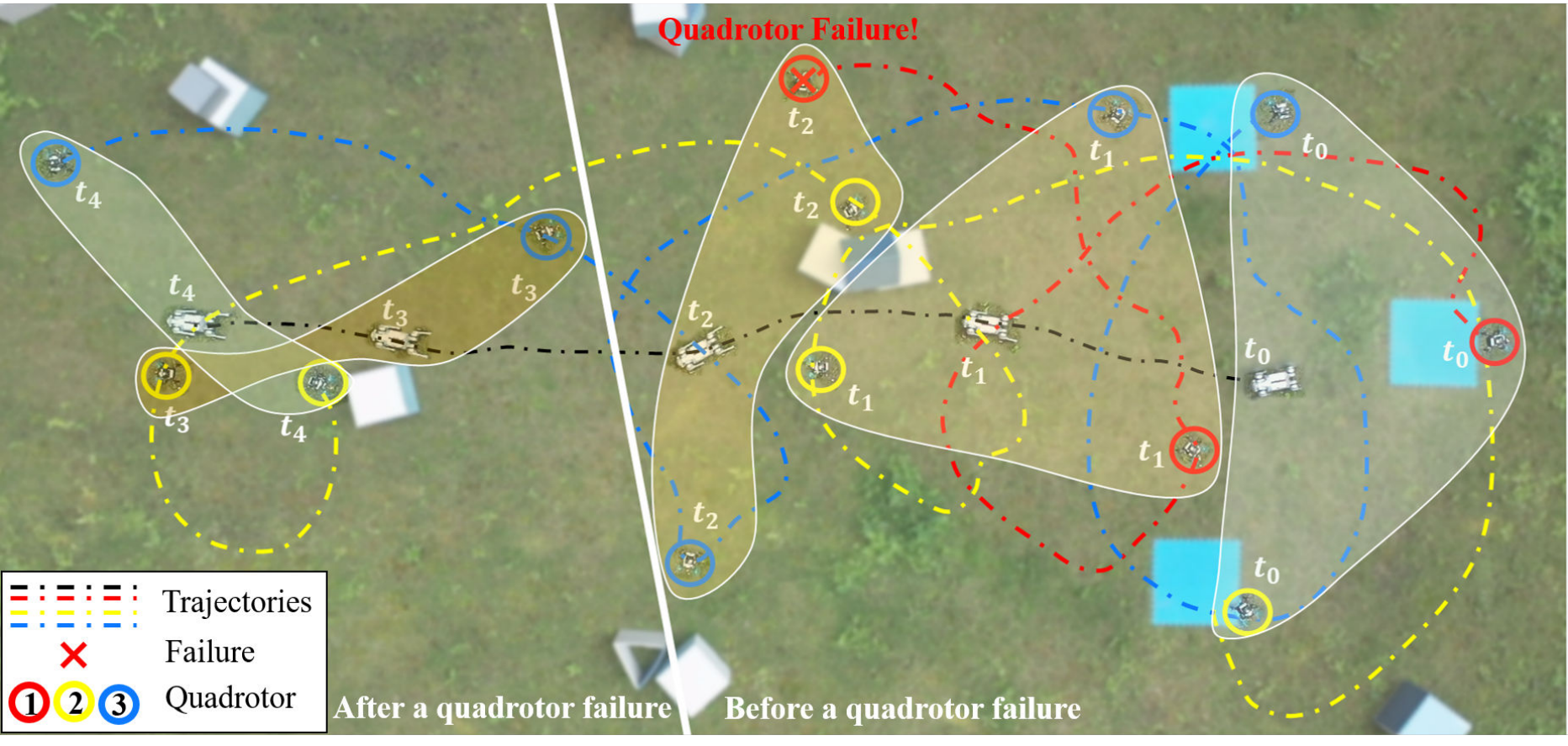}
        \caption{Outdoor experimental demonstration of three quadrotors cooperatively encircling a ground-moving target. The target traverses from the right to the left side of the image from $t_0$ to $t_4$. At $t_1$, the quadrotors form a triangular encirclement formation. Following a simulated quadrotor failure at $t_2$, the remaining two agents autonomously reconfigure their formation by $t_3$, achieving balanced positioning on opposite sides of the target.}
        \label{fig:outtraj}
    \end{figure}

The trajectories of both the quadrotors and the target are illustrated in Fig.~\ref{fig:outtraj}. During the initial mission phase, the quadrotors establish and maintain a triangular formation around the target ($t_1$ in Fig.~\ref{fig:outtraj}). As the formation progresses from $t_1$ to $t_2$, it successfully navigates through visually occluded regions while preserving complete encirclement (Fig.~\ref{fig:show}). The system demonstrates notable fault tolerance: when experiencing a simulated quadrotor failure at $t_2$, the remaining two agents autonomously reorganize into a stable symmetric configuration while maintaining uninterrupted target tracking ($t_3$ in Fig.~\ref{fig:outtraj}). These experimental outcomes conclusively verify the algorithm's dual competence in addressing both environmental occlusion constraints and unexpected agent failures during cooperative circumnavigation missions.

\section{Conclusion}
This paper presented a cooperative circumnavigation framework for multi-quadrotor systems operating in occluded environments using only onboard sensing. We developed distinct perception strategies and relative state estimation algorithms to facilitate cooperative interactions between different agent roles during missions. Experimental validation demonstrated that the proposed algorithm maintained robust performance even under quadrotor failures while dynamically adapting the circumnavigation formation in real time. Although the method was proved to be effective in outdoor environments, trajectory tracking and yaw angle control required further refinement. Future research directions include the integration of obstacle avoidance capabilities and extension to multi-target scenarios to improve the system's applicability to real-world search-and-rescue operations.

\section*{Acknowledgments}
This should be a simple paragraph before the References to thank those individuals and institutions who have supported your work on this article.

\bibliography{2025RAL} %

\begin{thebibliography}{10}
\providecommand{\url}[1]{#1}
\csname url@samestyle\endcsname
\providecommand{\newblock}{\relax}
\providecommand{\bibinfo}[2]{#2}
\providecommand{\BIBentrySTDinterwordspacing}{\spaceskip=0pt\relax}
\providecommand{\BIBentryALTinterwordstretchfactor}{4}
\providecommand{\BIBentryALTinterwordspacing}{\spaceskip=\fontdimen2\font plus
\BIBentryALTinterwordstretchfactor\fontdimen3\font minus
  \fontdimen4\font\relax}
\providecommand{\BIBforeignlanguage}[2]{{%
\expandafter\ifx\csname l@#1\endcsname\relax
\typeout{** WARNING: IEEEtran.bst: No hyphenation pattern has been}%
\typeout{** loaded for the language `#1'. Using the pattern for}%
\typeout{** the default language instead.}%
\else
\language=\csname l@#1\endcsname
\fi
#2}}
\providecommand{\BIBdecl}{\relax}
\BIBdecl

\bibitem{litimeinSurveyTechniquesCircular2021}
H.~Litimein, Z.-Y. Huang, and A.~Hamza, ``A {{Survey}} on {{Techniques}} in the
  {{Circular Formation}} of {{Multi-Agent Systems}},'' \emph{Electronics},
  vol.~10, no.~23, p. 2959, Nov. 2021.

\bibitem{deghatLocalizationCircumnavigationSlowly2014}
M.~Deghat, I.~Shames, B.~D.~O. Anderson, and C.~Yu, ``Localization and
  {{Circumnavigation}} of a {{Slowly Moving Target Using Bearing
  Measurements}},'' \emph{IEEE Trans. Automat. Contr.}, vol.~59, no.~8, pp.
  2182--2188, Aug. 2014.

\bibitem{shinzakiMultiAUVSystemCooperative2013}
D.~Shinzaki, C.~Gage, S.~Tang, M.~Moline, B.~Wolfe, C.~G. Lowe, and C.~Clark,
  ``A multi-{{AUV}} system for cooperative tracking and following of leopard
  sharks,'' in \emph{2013 {{IEEE International Conference}} on {{Robotics}} and
  {{Automation}}}.\hskip 1em plus 0.5em minus 0.4em\relax Karlsruhe, Germany:
  IEEE, May 2013, pp. 4153--4158.

\bibitem{liCooperativeCircumnavigationControl2020}
D.~Li, G.~Ma, W.~He, S.~S. Ge, and T.~H. Lee, ``Cooperative {{Circumnavigation
  Control}} of {{Networked Microsatellites}},'' \emph{IEEE Trans. Cybern.},
  vol.~50, no.~10, pp. 4550--4555, Oct. 2020.

\bibitem{mooreSourceSeekingCollaborative2010}
B.~J. Moore and C.~{Canudas-de-Wit}, ``Source seeking via collaborative
  measurements by a circular formation of agents,'' in \emph{Proceedings of the
  2010 {{American Control Conference}}}.\hskip 1em plus 0.5em minus 0.4em\relax
  Baltimore, MD: IEEE, Jun. 2010, pp. 6417--6422.

\bibitem{lopez-nicolasAdaptiveMultirobotFormation2020}
G.~{Lopez-Nicolas}, M.~Aranda, and Y.~Mezouar, ``Adaptive {{Multirobot
  Formation Planning}} to {{Enclose}} and {{Track}} a {{Target With Motion}}
  and {{Visibility Constraints}},'' \emph{IEEE Trans. Robot.}, vol.~36, no.~1,
  pp. 142--156, Feb. 2020.

\bibitem{suiAdaptiveBearingOnlyTarget2024}
D.~Sui, M.~Deghat, Z.~Sun, and M.~Eskandari, ``Adaptive {{Bearing-Only Target
  Localization}} and {{Circumnavigation Under Unknown Wind Disturbance}}:
  {{Theory}} and {{Experiments}},'' \emph{IEEE Robot. Autom. Lett.}, vol.~9,
  no.~12, pp. 11\,321--11\,328, Dec. 2024.

\bibitem{wangTargetTrackingCircumnavigation2024}
D.~Wang, Z.~Zhang, Y.~Zhao, and C.~Xu, ``Target {{Tracking With
  Circumnavigation Scheme Using Discrete Bearing}} and {{Control Input}},''
  \emph{IEEE Trans. Aerosp. Electron. Syst.}, vol.~60, no.~4, pp. 4699--4714,
  Aug. 2024.

\bibitem{chungSurveyAerialSwarm2018}
S.-J. Chung, A.~A. Paranjape, P.~Dames, S.~Shen, and V.~Kumar, ``A {{Survey}}
  on {{Aerial Swarm Robotics}},'' \emph{IEEE Trans. Robot.}, vol.~34, no.~4,
  pp. 837--855, Aug. 2018.

\bibitem{changCrossDroneBinocularCoordination2020}
Y.~Chang, H.~Zhou, X.~Wang, L.~Shen, and T.~Hu, ``Cross-{{Drone Binocular
  Coordination}} for {{Ground Moving Target Tracking}} in {{Occlusion-Rich
  Scenarios}},'' \emph{IEEE Robot. Autom. Lett.}, vol.~5, no.~2, pp.
  3161--3168, Apr. 2020.

\bibitem{zhouSwarmMicroFlying2022}
X.~Zhou, X.~Wen, Z.~Wang, Y.~Gao, H.~Li, Q.~Wang, T.~Yang, H.~Lu, Y.~Cao,
  C.~Xu, and F.~Gao, ``Swarm of micro flying robots in the wild,'' \emph{Sci.
  Robot.}, vol.~7, no.~66, p. eabm5954, May 2022.

\bibitem{ibenthalLocalizationPartiallyHidden2023}
J.~Ibenthal, L.~Meyer, H.~{Piet-Lahanier}, and M.~Kieffer, ``Localization of
  {{Partially Hidden Moving Targets Using}} a {{Fleet}} of {{UAVs}} via
  {{Bounded-Error Estimation}},'' \emph{IEEE Trans. Robot.}, vol.~39, no.~6,
  pp. 4211--4229, Dec. 2023.

\bibitem{liuFormationControlEnclosing2025}
X.~Liu, D.~Zhang, Q.~Zhang, and T.~Hu, ``Formation {{Control}} for
  {{Enclosing}} and {{Tracking}} via {{Relative Localization}},'' no.
  arXiv:2410.14407, Feb. 2025.

\bibitem{hausmanCooperativeMultirobotControl2015}
K.~Hausman, J.~M{\"u}ller, A.~Hariharan, N.~Ayanian, and G.~S. Sukhatme,
  ``Cooperative multi-robot control for target tracking with onboard sensing,''
  \emph{The International Journal of Robotics Research}, vol.~34, no.~13, pp.
  1660--1677, Nov. 2015.

\bibitem{priceDeepNeuralNetworkBased2018}
E.~Price, G.~Lawless, R.~Ludwig, I.~Martinovic, H.~H. Bulthoff, M.~J. Black,
  and A.~Ahmad, ``Deep {{Neural Network-Based Cooperative Visual Tracking
  Through Multiple Micro Aerial Vehicles}},'' \emph{IEEE Robot. Autom. Lett.},
  vol.~3, no.~4, pp. 3193--3200, Oct. 2018.

\bibitem{xuDistributedPseudolinearEstimation2017}
S.~Xu, K.~Do{\u g}an{\c c}ay, and H.~Hmam, ``Distributed pseudolinear
  estimation and {{UAV}} path optimization for {{3D AOA}} target tracking,''
  \emph{Signal Processing}, vol. 133, pp. 64--78, Apr. 2017.

\bibitem{zhangRobustNonlinearClose2021}
Q.~Zhang and H.~H.~T. Liu, ``Robust {{Nonlinear Close Formation Control}} of
  {{Multiple Fixed-Wing Aircraft}},'' \emph{Journal of Guidance, Control, and
  Dynamics}, vol.~44, no.~3, pp. 572--586, Mar. 2021.

\bibitem{GBZhu_T-Mech2024}
G.~Zhu, Q.~Zhang, B.~Zhu, and T.~Hu, ``Heuristic predictive control for
  multirobot flocking in congested environments,'' \emph{IEEE/ASME Transactions
  on Mechatronics}, pp. 1--12, Sep. 2024, (Early Access).

\bibitem{liFullyDistributedCooperative2021}
D.~Li, K.~Cao, L.~Kong, and H.~Yu, ``Fully {{Distributed Cooperative
  Circumnavigation}} of {{Networked Unmanned Aerial Vehicles}},''
  \emph{IEEE/ASME Trans. Mechatron.}, vol.~26, no.~2, pp. 709--718, Apr. 2021.

\bibitem{zhangCompositeSystemTheorybased2021}
M.~Zhang, J.~Jia, and J.~Mei, ``A composite system theory-based guidance law
  for cooperative target circumnavigation of {{UAVs}},'' \emph{Aerospace
  Science and Technology}, vol. 118, p. 107034, Nov. 2021.

\bibitem{zhangRobustGuidanceLaw2023}
M.~Zhang, C.~Liang, and J.~Mei, ``Robust guidance law for cooperative aerial
  target circumnavigation of {{UAVs}} based on composite system theory,''
  \emph{Aerospace Science and Technology}, vol. 140, p. 108439, Sep. 2023.

\bibitem{chenFormationCircumnavigationUnmanned2019}
Y.-Y. Chen, Y.~Zhang, C.-L. Liu, and Q.~Wang, ``Formation circumnavigation for
  unmanned aerial vehicles using relative measurements with an uncertain
  dynamic target,'' \emph{Nonlinear Dyn}, vol.~97, no.~4, pp. 2305--2321, Sep.
  2019.

\bibitem{liVGSwarmVisionBasedGene2023}
H.~Li, Y.~Cai, J.~Hong, P.~Xu, H.~Cheng, X.~Zhu, B.~Hu, Z.~Hao, and Z.~Fan,
  ``{{VG-Swarm}}: {{A Vision-Based Gene Regulation Network}} for {{UAVs Swarm
  Behavior Emergence}},'' \emph{IEEE Robot. Autom. Lett.}, vol.~8, no.~3, pp.
  1175--1182, Mar. 2023.

\bibitem{zhengEnclosingTargetNonholonomic2015}
R.~Zheng, Y.~Liu, and D.~Sun, ``Enclosing a target by nonholonomic mobile
  robots with bearing-only measurements,'' \emph{Automatica}, vol.~53, pp.
  400--407, Mar. 2015.

\bibitem{liLocalizationCircumnavigationMultiple2018}
R.~Li, Y.~Shi, and Y.~Song, ``Localization and circumnavigation of multiple
  agents along an unknown target based on bearing-only measurement: {{A}} three
  dimensional solution,'' \emph{Automatica}, vol.~94, pp. 18--25, Aug. 2018.

\bibitem{zouCoordinateFreeDistributedLocalization2022}
Y.~Zou, W.~Yang, W.~He, Q.~Fu, Q.~Li, and C.~Silvestre, ``Coordinate-{{Free
  Distributed Localization}} and {{Circumnavigation}} for {{Nonholonomic
  Vehicles Without Position Information}},'' \emph{IEEE/ASME Trans.
  Mechatron.}, vol.~27, no.~5, pp. 2523--2534, Oct. 2022.

\bibitem{hungCooperativeDistributedEstimation2022}
N.~T. Hung, F.~F.~C. Rego, and A.~M. Pascoal, ``Cooperative {{Distributed
  Estimation}} and {{Control}} of {{Multiple Autonomous Vehicles}} for
  {{Range-Based Underwater Target Localization}} and {{Pursuit}},'' \emph{IEEE
  Trans. Contr. Syst. Technol.}, vol.~30, no.~4, pp. 1433--1447, Jul. 2022.

\bibitem{liuMovingTargetCircumnavigationUsing2023}
F.~Liu, C.~Guo, W.~Meng, R.~Su, and H.~Li, ``Moving-{{Target Circumnavigation
  Using Adaptive Neural Anti-Synchronization Control}} via {{Distance-Only
  Measurements}},'' \emph{IEEE Trans. Cybern.}, pp. 1--11, 2023.

\bibitem{liuFormationControlMoving2023}
X.~Liu, K.~Liu, T.~Hu, and Q.~Zhang, ``Formation {{Control}} for {{Moving
  Target Enclosing}} via {{Relative Localization}},'' in \emph{2023 62nd {{IEEE
  Conference}} on {{Decision}} and {{Control}} ({{CDC}})}.\hskip 1em plus 0.5em
  minus 0.4em\relax Singapore, Singapore: IEEE, Dec. 2023, pp. 1400--1405.

\bibitem{zouLeaderFollowerCircumnavigation2024}
Y.~Zou, L.~Zhong, W.~He, and C.~Silvestre, ``Leader--follower circumnavigation
  control of non-holonomic robots using distance-related information,''
  \emph{Automatica}, vol. 169, p. 111831, Nov. 2024.

\bibitem{nielsenRelativeMovingTarget2019}
J.~Nielsen and R.~Beard, ``Relative {{Moving Target Tracking}} and
  {{Circumnavigation}},'' in \emph{2019 {{American Control Conference}}
  ({{ACC}})}.\hskip 1em plus 0.5em minus 0.4em\relax Philadelphia, PA, USA:
  IEEE, Jul. 2019, pp. 1122--1127.

\bibitem{ningRealtoSimtoRealApproachVisionBased2024}
Z.~Ning, Y.~Zhang, X.~Lin, and S.~Zhao, ``A {{Real-to-Sim-to-Real Approach}}
  for {{Vision-Based Autonomous MAV-Catching-MAV}},'' \emph{Un. Sys.}, vol.~12,
  no.~04, pp. 787--798, Jul. 2024.

\bibitem{Zhang2015RAL}
D.~Zhang, C.~Yu, F.~Xue, and Q.~Zhang, ``Learning efficient flocking control
  based on gibbs random fields,'' \emph{IEEE Robotics and Automation Letters},
  vol.~10, no.~4, pp. 3478--3485, 2025.

\bibitem{nguyenDistanceBasedCooperativeRelative2019}
T.-M. Nguyen, Z.~Qiu, T.~H. Nguyen, M.~Cao, and L.~Xie, ``Distance-{{Based
  Cooperative Relative Localization}} for {{Leader-Following Control}} of
  {{MAVs}},'' \emph{IEEE Robot. Autom. Lett.}, vol.~4, no.~4, pp. 3641--3648,
  Oct. 2019.

\bibitem{nguyenPersistentlyExcitedAdaptive2020}
------, ``Persistently {{Excited Adaptive Relative Localization}} and
  {{Time-Varying Formation}} of {{Robot Swarms}},'' \emph{IEEE Trans. Robot.},
  vol.~36, no.~2, pp. 553--560, Apr. 2020.

\bibitem{guoUltraWidebandOdometryBasedCooperative2020}
K.~Guo, X.~Li, and L.~Xie, ``Ultra-{{Wideband}} and {{Odometry-Based
  Cooperative Relative Localization With Application}} to {{Multi-UAV Formation
  Control}},'' \emph{IEEE Trans. Cybern.}, vol.~50, no.~6, pp. 2590--2603, Jun.
  2020.

\bibitem{cossetteRelativePositionEstimation2021}
C.~C. Cossette, M.~Shalaby, D.~Saussie, J.~R. Forbes, and J.~Le~Ny, ``Relative
  {{Position Estimation Between Two UWB Devices With IMUs}},'' \emph{IEEE
  Robot. Autom. Lett.}, vol.~6, no.~3, pp. 4313--4320, Jul. 2021.

\bibitem{zhangAgileFormationControl2022}
P.~Zhang, G.~Chen, Y.~Li, and W.~Dong, ``Agile {{Formation Control}} of {{Drone
  Flocking Enhanced With Active Vision-Based Relative Localization}},''
  \emph{IEEE Robot. Autom. Lett.}, vol.~7, no.~3, pp. 6359--6366, Jul. 2022.

\bibitem{zhengUWBVIOFusionAccurate2022}
S.~Zheng, Z.~Li, Y.~Liu, H.~Zhang, P.~Zheng, X.~Liang, Y.~Li, X.~Bu, and
  X.~Zou, ``{{UWB-VIO Fusion}} for {{Accurate}} and {{Robust Relative
  Localization}} of {{Round Robotic Teams}},'' \emph{IEEE Robot. Autom. Lett.},
  vol.~7, no.~4, pp. 11\,950--11\,957, Oct. 2022.

\bibitem{fishbergMURPMultiAgentUltraWideband2024}
A.~Fishberg, B.~Quiter, and J.~P. How, ``{{MURP}}: {{Multi-Agent Ultra-Wideband
  Relative Pose Estimation}} with {{Constrained Communications}} in {{3D
  Environments}},'' \emph{IEEE Robot. Autom. Lett.}, pp. 1--8, 2024.

\bibitem{wangCooperativeTargetTracking2012}
Z.~Wang and D.~Gu, ``Cooperative {{Target Tracking Control}} of {{Multiple
  Robots}},'' \emph{IEEE Trans. Ind. Electron.}, vol.~59, no.~8, pp.
  3232--3240, Aug. 2012.

\bibitem{olfati-saberCollaborativeTargetTracking2011}
R.~{Olfati-Saber} and P.~Jalalkamali, ``Collaborative target tracking using
  distributed {{Kalman}} filtering on mobile sensor networks,'' in
  \emph{Proceedings of the 2011 {{American Control Conference}}}.\hskip 1em
  plus 0.5em minus 0.4em\relax San Francisco, CA: IEEE, Jun. 2011, pp.
  1100--1105.

\bibitem{liDistributedKalmanFilter2020}
W.~Li, Y.~Jia, and J.~Du, ``Distributed {{Kalman Filter}} for {{Cooperative
  Localization With Integrated Measurements}},'' \emph{IEEE Trans. Aerosp.
  Electron. Syst.}, vol.~56, no.~4, pp. 3302--3310, Aug. 2020.

\bibitem{doostmohammadianDistributedEstimationApproach2022}
M.~Doostmohammadian, A.~Taghieh, and H.~Zarrabi, ``Distributed {{Estimation
  Approach}} for {{Tracking}} a {{Mobile Target}} via {{Formation}} of
  {{UAVs}},'' \emph{IEEE Trans. Automat. Sci. Eng.}, vol.~19, no.~4, pp.
  3765--3776, Oct. 2022.

\bibitem{dingOverviewRecentAdvances2018}
L.~Ding, Q.-L. Han, X.~Ge, and X.-M. Zhang, ``An {{Overview}} of {{Recent
  Advances}} in {{Event-Triggered Consensus}} of {{Multiagent Systems}},''
  \emph{IEEE Trans. Cybern.}, vol.~48, no.~4, pp. 1110--1123, Apr. 2018.

\bibitem{prielEventtriggeredConsensusKalman2023}
A.~Priel and D.~Zelazo, ``Event-triggered consensus {{Kalman}} filtering for
  time-varying networks and intermittent observations,'' \emph{Intl J Robust \&
  Nonlinear}, vol.~33, no.~13, pp. 7430--7451, Sep. 2023.

\bibitem{xuOmniSwarmDecentralizedOmnidirectional2022}
H.~Xu, Y.~Zhang, B.~Zhou, L.~Wang, X.~Yao, G.~Meng, and S.~Shen,
  ``Omni-{{Swarm}}: {{A Decentralized Omnidirectional
  Visual}}--{{Inertial}}--{{UWB State Estimation System}} for {{Aerial
  Swarms}},'' \emph{IEEE Trans. Robot.}, vol.~38, no.~6, pp. 3374--3394, Dec.
  2022.

\bibitem{Jocher_YOLOv5_by_Ultralytics_2020}
\BIBentryALTinterwordspacing
G.~Jocher, ``{YOLOv5 by Ultralytics},'' May 2020. [Online]. Available:
  \url{https://github.com/ultralytics/yolov5}
\BIBentrySTDinterwordspacing

\bibitem{mellingerMinimumSnapTrajectory2011}
D.~Mellinger and V.~Kumar, ``Minimum snap trajectory generation and control for
  quadrotors,'' in \emph{2011 {{IEEE International Conference}} on {{Robotics}}
  and {{Automation}}}.\hskip 1em plus 0.5em minus 0.4em\relax Shanghai, China:
  IEEE, May 2011, pp. 2520--2525.

\end{thebibliography}
\bibliographystyle{IEEEtran}

\end{document}